%% file: main.tex
\documentclass[conference]{IEEEtran}
\IEEEoverridecommandlockouts
\usepackage{cite}
\usepackage{subcaption} 
\usepackage{amsmath,amssymb,amsfonts}
\usepackage{algorithmic}
\usepackage{graphicx}
\usepackage{textcomp}
\usepackage{algorithm}
\usepackage{soul}
\usepackage{mathtools}
\usepackage{tabularx}
\usepackage{booktabs}
\usepackage{threeparttable}
\usepackage{hyperref}
\usepackage[table, dvipsnames]{xcolor}

\usepackage[super]{nth}
\usepackage{siunitx}
\usepackage{standalone}
\usepackage{tikz}
\usepackage{multirow} 
\usepackage{hhline}
\usepackage{multicol} 
\usepackage{bm}
\definecolor{headerblue}{HTML}{0F5E7A}
\definecolor{stripe}{HTML}{F3F5F7}
\definecolor{hilite}{HTML}{FFF2A8}

\def\BibTeX{{\rm B\kern-.05em{\sc i\kern-.025em b}\kern-.08em
    T\kern-.1667em\lower.7ex\hbox{E}\kern-.125emX}}
\begin{document}

\title{Delay-Aware Reinforcement Learning for Highway On-Ramp Merging under Stochastic Communication Latency}
\author{
Amin Tabrizian\textsuperscript{\textdagger}, 
Zhitong Huang\textsuperscript{\textdaggerdbl}, 
Arsyi Aziz\textsuperscript{\textdagger}, 
Peng Wei\textsuperscript{\textsection} \\
\thanks{\textsuperscript{\textdagger}Department of Computer Science, George Washington University, Washington, D.C. (email: amin\_tabrizian@gwu.edu)}%
\thanks{\textsuperscript{\textdaggerdbl}Connected and Automated Vehicle Program Manager, Traffic Operations Division, Virginia Department of Transportation.}%
\thanks{\textsuperscript{\textsection}Department of Mechanical \& Aerospace Engineering, George Washington University, Washington, D.C.}%
\thanks{This work is supported by National Science Foundation Award \#2229885.}%
\thanks{Copyright (c) 2026 IEEE. Personal use of this material is permitted. However, permission to use this material for any other purposes must be obtained from the IEEE by sending a request to pubs-permissions\@ieee.org.}
}

\maketitle
\begin{abstract}
Delayed and partially observable state information poses significant challenges for reinforcement learning (RL)-based control in real-world autonomous driving. In highway on-ramp merging, a roadside unit (RSU) can sense nearby traffic, perform edge perception, and transmit state estimates to the ego vehicle over vehicle-to-infrastructure (V2I) links. With recent advancements in intelligent transportation infrastructure and edge computing, such RSU-assisted perception is increasingly realistic and already deployed in modern connected roadway systems. However, edge processing time and wireless transmission can introduce stochastic V2I communication delays, violating the Markov assumption and substantially degrading control performance. In this work, we propose \textbf{DAROM}, a \textit{D}elay-\textit{A}ware \textit{R}einforcement Learning framework for \textit{O}n-ramp \textit{M}erging that is robust to stochastic delays. We model the problem as a random delay Markov decision process (RDMDP) and develop a unified RL agent for joint longitudinal and lateral control. To recover a Markovian representation under delayed observations, we introduce a \textbf{Delay-Aware Encoder} that conditions on delayed observations, masked action histories, and observed delay magnitude to infer the current latent state. We further integrate a physics-based safety controller to reduce collision risk during merging. Experiments in the simulation of urban mobility (SUMO) simulator using real-world traffic data from the Next Generation Simulation (NGSIM) dataset demonstrate that \textbf{DAROM} consistently outperforms standard RL baselines across traffic densities. In particular, the gated recurrent unit (GRU)-based encoder achieves over 99\% success in high-density traffic with random V2I delays of up to 2.0 seconds.
\end{abstract}


\section{Introduction} \label{sec:intro}
\input{01-Introduction/intro}

\section{Preliminaries} \label{sec:prelim}
\input{02-Preliminaries/prelim}
\section{Problem Formulation} \label{sec:formulation}
\input{03-ProblemFormulation/formulation}
\section{Solution Method} \label{sec:method}
\input{04-Method/method}

\section{Experiments} \label{sec:Experiments}
\input{05-Experiments/experiments}
\section{Conclusion and Future Work} \label{sec:Conclusion}
\input{06-Conclusion/conclusion}
\bibliographystyle{support/IEEEtran}
\bibliography{support/ref}

\end{document}

%% file: 01-Introduction/intro.tex
\subsection{Motivation and Related Work}
In modern transportation systems, autonomous driving has emerged as a promising technology for improving road safety, traffic efficiency, and driving comfort by reducing human error and enabling more consistent vehicle behavior \cite{FAGNANT2015167}. A key challenge in realizing these benefits lies in the safe and efficient integration of autonomous vehicles into complex and highly interactive traffic scenarios, such as highway on-ramp merging \cite{7562449}. Highway on-ramp merging areas are widely recognized as traffic bottlenecks and safety-critical regions due to their limited roadway capacity, high traffic density, and frequent lane-changing and speed-adjustment maneuvers performed by merging and mainline vehicles \cite{FHWA2017Bottlenecks}.

With the recent successes of reinforcement learning (RL) in board games \cite{silver2018general} and Atari \cite{mnih2013playing}, researchers applied RL-based algorithms to extensive real-world applications such as natural language processing \cite{sharma2017literature}, aviation \cite{razzaghi2022survey}, and autonomous driving \cite{wang2017formulation}. In the case of highway on-ramp merging, numerous studies leverage RL techniques to ensure a safe and effective merging. An on-ramp merging optimal control framework based on a deep RL that optimizes lane keeping and lane-changing at the same time was proposed in \cite{ZHAO2020542}. The proposed approach had a shorter travel time and lower emergency braking rate compared to baseline methods. RL and model predictive control (MPC) were combined in \cite{lubars2021combining} to leverage their strengths together. The authors claimed that MPC solutions provide more robustness to out-of-distribution traffic patterns and RL techniques are better in terms of efficiency and passenger comfort. They considered a single-lane highway on-ramp merging scenario with a default driving style for the surrounding vehicles.

Developing safe merging control algorithms was the aim of some other studies. A real-time bi-level control framework was proposed in \cite{lyu2021probabilistic} that ensures safety using a control barrier function (CBF). Combining learning-based control methods with CBF was not addressed in their work. CBFs always require a good approximation of the system model, which may not be applicable to all driving scenarios. A probabilistic CBF algorithm was proposed in \cite{PCBF} to account for model uncertainty. The proposed algorithm was tested based on the Next Generation Simulation (NGSIM) provided by the Federal Highway Administration (FHWA) \cite{ngsim}.

Instead of using direct vehicle-to-vehicle (V2V) communication, \cite{9197430} focused on utilizing realistic sensor data mounted on the ego vehicle to obtain observations as input for its control framework. They employed a game-theoretic reasoning based on Monte Carlo RL in order to find a near-optimal policy. A passive actor-critic (pAC) technique was suggested in \cite{8671752} for selecting a candidate spot for merging and then used a multi-policy decision-making method to merge to candidate spots. They then validated their proposed framework on real-world data with a success rate of $92\%$.

Latent state inference plays an important role in autonomous driving. Knowing surrounding vehicles' intent to yield or not is crucial for ego vehicle's planning and control. The drivers' behavior or intention on other vehicles can be estimated by reasoning their underlying driving styles. A supervised learning approach for inferring the latent states of other vehicles in a T-intersection was proposed in \cite{9562006}. They also adopted graph neural networks (GNNs) to model relational information of the neighboring vehicles.

Despite the substantial progress made by prior work, most existing on-ramp merging approaches implicitly assume timely and locally available state information, obtained either through idealized full-state access, onboard sensing, or direct V2V communication. In contrast, our setting explicitly considers an infrastructure-assisted perception pipeline, where the ego vehicle relies on roadside unit (RSU)-derived state estimates that are subject to stochastic processing and communication delays. While several studies leverage RL, MPC, or safety filters for merging control, they typically formulate the problem as a standard Markov decision process or partially observable MDP (POMDP) without accounting for random observation latency induced by vehicle-to-infrastructure (V2I) communication. Moreover, existing works on latent state inference focus primarily on modeling driver intent or interaction structure, rather than recovering the Markov property under delayed observations. As a result, the impact of stochastic V2I latency on decision-making performance in safety-critical merging scenarios remains largely unexplored. This gap motivates our delay-aware formulation and learning framework, which is specifically designed to operate under infrastructure-based perception with random communication delays.

\subsection{RSU-Based Perception} 

RSUs are fixed infrastructure components deployed along highways and urban roads to support V2I communication and perception. Beyond simple message relaying, modern RSUs are increasingly equipped with infrastructure-grade sensors such as cameras, light detection and ranging (LiDAR), and radar, along with edge computing capabilities that enable real-time perception of the traffic environment. Recent advances in infrastructure-based cooperative perception demonstrate that RSUs can reliably estimate surrounding vehicles’ positions, velocities, and behaviors using computer vision and sensor fusion techniques, and transmit this processed information to connected vehicles via V2I links \cite{Yu_2022_CVPR}, \cite{s20185320}.

Large-scale datasets and real-world deployments further validate the feasibility of this paradigm. The DAIR-V2X dataset provides synchronized camera and LiDAR data collected from roadside infrastructure, showing that accurate 3D object detection and tracking can be achieved solely from infrastructure sensors and shared with vehicles \cite{Yu_2022_CVPR}. More recent work has demonstrated communication-efficient collaborative perception frameworks that explicitly account for bandwidth and latency constraints when transmitting infrastructure-derived perception results to vehicles \cite{feng2025lcv2icommunicationefficienthighperformancecollaborative}. Prototype systems integrating cameras, radar, and vehicle-to-everything (V2X) communication within a single RSU have also been deployed in real traffic environments, enabling real-time object detection, tracking, and behavior prediction \cite{adas,rsuv2x}.

In this work, we assume that the ego vehicle does not rely on V2V communication and is not equipped with onboard sensors for estimating the states of surrounding vehicles. Instead, an RSU captures traffic information in the merging region and communicates it to the ego vehicle via V2I links. This modeling choice is motivated by the geometric and environmental characteristics of highway on-ramp merging scenarios, where curved road layouts, overpasses, shoulders, and dense traffic frequently obstruct direct line-of-sight between vehicles. Under such conditions, peer-to-peer sensing or communication alone cannot ensure reliable or complete situational awareness. In contrast, infrastructure-based perception benefits from an elevated and fixed vantage point, enabling continuous monitoring of the merging area and aggregation of observations over space and time~\cite{s20185320}.

\subsection{Stochastic V2I Communication Latency}
Despite the advantages of V2I communication, it introduces non-negligible and stochastic latency. End-to-end delay arises from multiple sources, including edge processing time, wireless transmission, packet scheduling, and queuing within network stacks. Field experiments with networked roadside perception units have shown that, even under worst-case conditions, processed perception messages can typically be delivered to vehicles within approximately 100\,ms; however, the observed delay varies depending on traffic density and communication load \cite{s20185320}. When combined with additional processing and networking variability, these effects motivate modeling V2I communication latency as a stochastic process rather than a fixed constant.

Highway on-ramp merging represents a particularly safety-critical application for infrastructure-based perception under delayed observations. The limited acceleration lane length, high relative speeds, and frequent gap acceptance decisions make timely awareness of surrounding traffic essential. These characteristics amplify the impact of delayed information on decision making. Accordingly, this paper focuses on highway on-ramp merging as a representative and challenging scenario, assuming RSU-based perception with stochastic V2I communication latency and no reliance on V2V communication. Limitations related to packet loss, cooperative multi-agent communication, and tighter latency guarantees are left for future work.


\subsection{Contributions}

To the best of our knowledge, no existing study has investigated stochastic V2I communication latency within the autonomous highway on-ramp merging setting. To bridge this gap, we introduce \textbf{DAROM}, a delay-aware reinforcement learning framework for on-ramp merging designed to enable the ego vehicle to make robust acceleration and lane-changing decisions in highway on-ramp merging scenarios, even under stochastic communication delays. The system integrates multiple components to enhance driving safety and robustness in practical applications:

\begin{enumerate}
    \item \textbf{Unified RL Agent}: A Deep RL (DRL) agent that jointly optimizes longitudinal acceleration and lateral lane-changing maneuvers. Unlike hierarchical approaches, this unified agent outputs continuous control commands for both dimensions simultaneously. 
    
    \item \textbf{Delay-Aware Encoder}: A specialized neural network module designed to handle the RDMDP. It processes the augmented state, comprising the delayed observation, the masked history of past actions, and the current delay magnitude, to implicitly infer the real-time latent state of the traffic environment.
    
    \item \textbf{Physics-Based Safety Controller}: A supervisory control layer deployed to ensure the safety of the learning-based agent. It evaluates the agent's proposed actions against kinematic stopping distance constraints, overriding maneuvers that violate safety gaps.
\end{enumerate}
Our approach explicitly addresses the challenge of stochastic latency in infrastructure-based perception, ensuring robustness where standard RL methods fail.

The main contributions of our work are:
\begin{itemize}
    \item \textbf{Unified Control Architecture:} We propose a single DRL agent capable of simultaneously executing lane-keeping and lane-changing decisions, replacing complex hierarchical structures with an end-to-end learning objective.
    \item \textbf{RDMDP Modeling:} We model the random delay MDP for highway on-ramp merging to recover the Markov property through state augmentation and introduce a Delay-Aware Encoder that effectively handles the stochastic communication delays induced by the RSU.
    \item \textbf{Delay-Aware Encoder:} We design a specialized encoder that processes delayed observations, masked histories of past actions, and the delay magnitude to implicitly infer the current latent traffic state.
    \item \textbf{Robust Evaluation:} We construct extensive simulations based on real-world traffic data to demonstrate the framework's superior performance and stability under stochastic delay conditions compared to state-of-the-art baselines.
\end{itemize}

The remainder of this paper is organized as follows: Section \ref{sec:prelim} establishes the preliminary background. Section \ref{sec:formulation} outlines the problem formulation, specifying the state space, action space, and reward structure for the on-ramp merging scenario. Section \ref{sec:method} details the proposed DAROM framework, including the DRL agent, the specific architecture of the Delay-Aware Encoder, and the physics-based safety controller. Section \ref{sec:Experiments} presents the experimental setup, comparative baselines, and a comprehensive discussion of the simulation results. Finally, Section \ref{sec:Conclusion} concludes the paper and discusses directions for future research.

%% file: 02-Preliminaries/prelim.tex
\subsection{Standard Markov Decision Processes}
We consider a sequential decision-making problem modeled as an MDP, defined by the tuple $(\mathcal{S}, \mathcal{A}, P, r, \gamma)$. At each time-step $t$, the agent observes a state $s_t \in \mathcal{S}$, executes an action $a_t \in \mathcal{A}$, and receives a reward $r(s_t, a_t)$. The environment evolves according to the transition probability $P(s_{t+1} | s_t, a_t)$, and the goal is to maximize the expected discounted cumulative return $\mathbb{E}[\sum_{k=0}^\infty \gamma^k r_{t+k}]$. Here $\gamma \in [0, 1)$ is the discount factor.

\subsection{Random Delay Markov Decision Processes}
In our specific setting, the agent does not observe the true state $s_t$ immediately due to processing and transmission latencies at the RSU. We model this using the RDMDP framework proposed by \cite{bouteiller2021reinforcement}.

Unlike the general formulation in \cite{bouteiller2021reinforcement}, our system is characterized by \textit{random observation delays} but no \textit{action delays}. The agent receives an observation $o_t$ corresponding to a past state $s_{t-\omega_t}$, where $\omega_t$ is the delay of the received observation at time-step $t$. The action $a_t$ is applied immediately to the environment.

We adapt Definition 1 from \cite{bouteiller2021reinforcement} to our setting as follows:

\textbf{Definition 1.} An RDMDP is defined as a tuple $(\mathcal{X}, \mathcal{A}, \tilde{\mu}, \bar{p})$, which augments an underlying MDP $E=(\mathcal{S}, \mathcal{A}, P, r, \gamma)$ with initial state distribution $\mu$ as follows:
\begin{enumerate}
    \item \textbf{Augmented State-Space} $\mathcal{X} = \mathcal{S} \times \mathcal{A}^{K} \times \mathbb{N}$. A state $x \in \mathcal{X}$ is a tuple $(o, u, \omega)$, where:
    \begin{itemize}
        \item $o \in \mathcal{S}$ is the \textit{delayed observation} (i.e., $s_{t-\omega}$).
        \item $u \in \mathcal{A}^{K}$ is a buffer of the last $K$ sent actions. $K$ represents the fixed maximum possible delay.
        \item $\omega \in \mathbb{N}$ is the delay of the received observation.
    \end{itemize}
    \item \textbf{Action-Space} $\mathcal{A}$, identical to the underlying MDP.
    \item \textbf{Initial State Distribution} $\tilde{\mu}(x_0) = \tilde{\mu}(o, u, \omega) = \mu(o)\delta(u-c_u)\delta(\omega-c_\omega)$, where $c_u$ and $c_\omega$ are initialization constants.
    \item \textbf{Transition Distribution} $\bar{p}$:
    \begin{equation}
        \begin{split}
            \bar{p}(o', u', \omega', r' & \mid o, u, \omega, a) = \\
            & f_{\omega - \omega'}\bigl(o', r' \mid o, u, \omega, a\bigr) \\
            & \cdot p_\omega(\omega'\mid\omega) \, p_u(u'\mid u, a)
        \end{split}
    \end{equation}
    
\end{enumerate}

Here, $p_\omega$ models the stochastic evolution of the delay, and $p_u$ describes the deterministic update of the action buffer:
\begin{equation}
    p_u(u'|u, a) = \delta(u' - (a, u[1:-1]))
\end{equation}
where the notation $(a, u[1:-1])$ represents shifting the buffer to include the new action $a$ at the head ($u'[0]$) and discarding the oldest action ($u[K]$). In this notation $f_{\omega - \omega'}$ describes the evolution of observations and rewards. This augmentation ensures the Markov property is preserved despite the delay, as the buffer $u$ captures the sequence of actions that have influenced the environment state since the delayed state $s_{t-\omega_t}$ was observed ($o_t$).

\begin{figure}[!h]
  \centering
  \includestandalone[width=\columnwidth]{Figures/highway_rsu}
  \caption{Highway on-ramp merging scenario. The RSU will compute the surrounding vehicles' information and communicates it to the ego vehicle (red) with stochastic latency.}
  \label{scenario}
\end{figure}

%% file: 03-ProblemFormulation/formulation.tex
\subsection{Scenario Description}
We apply the RDMDP framework to an autonomous highway on-ramp merging scenario, illustrated in Figure \ref{scenario}. The ego vehicle (red) starts on an acceleration lane and must merge into a dense flow of mixed traffic on the mainline. Surrounding vehicles are perceived by an RSU which transmits this information to the ego vehicle with stochastic latency, resulting in delayed observations. The environment is therefore partially observable due to the RSU-based perception delays described in Section~\ref{sec:intro}.

\subsection{Vehicle State and Action Definitions}\label{sec:formulation:state_action}
To ground the abstract RDMDP definitions, we specify the physical components of the underlying MDP.

\subsubsection{Underlying State Space ($\mathcal{S}$)}
The physical state $s \in \mathcal{S}$ encapsulates the kinematics of the ego vehicle and its nearest surrounding vehicles within a specified radius. For any surrounding vehicle $i$, the feature vector is defined as
\[
\phi^i = \big[p_{rel}^i,\; y_{rel}^i,\; v_{x,rel}^i\big],
\]
representing the longitudinal and lateral relative positions and the relative longitudinal velocity with respect to the ego vehicle. The ego feature vector $\phi^{\text{ego}}$ captures the ego vehicle’s corresponding kinematic state in the absolute frame.

The global state $s$ is constructed by concatenating the ego feature vector with the feature vectors of surrounding vehicles, which are sorted based on their relative lane position and longitudinal distance to the ego vehicle. To maintain a fixed-dimensional state representation, if fewer surrounding vehicles are present than the maximum supported by the observation encoder, the remaining entries are zero-padded.

We assume that surrounding vehicle information is collected by RSUs deployed along the highway and communicated to the ego vehicle. The ego vehicle’s on-board processing unit filters out vehicles beyond a predefined distance threshold before forming the state.

\subsubsection{Action Space ($\mathcal{A}$)}
The agent generates a continuous control vector $a \in \mathbb{R}^2$, which is mapped to the executable simulation commands $a = [a_{lk}, a_{lc}]^T$, where $a_{lk}$ is the longitudinal acceleration and $a_{lc}$ is the lane changing decision. 

\subsection{Reward Function}
To ensure the agent learns a policy that balances task completion, driving efficiency, passenger comfort, and safety, we design a comprehensive composite reward function. The total reward $R_t$ at time-step $t$ is computed as a scaled sum of distinct components:
\begin{equation}
    R_t =   r_\text{step} + r_\text{prog} + r_\text{comf} + r_\text{safe} + r_\text{event}
\end{equation}
The individual terms are defined as follows:

\subsubsection{Continuous Driving Rewards}
The agent is incentivized to maintain velocity and smooth control while minimizing the time spent in the merging zone:
\begin{equation}
    r_\text{step} = -c_\text{exist}
\end{equation}
\begin{equation}
    r_\text{prog} = \lambda_\text{prog} \cdot (p_{t}^{\text{ego}} - p_{t-1}^{\text{ego}})
\end{equation}
\begin{equation}
    r_\text{comf} = -\lambda_\text{jerk} \cdot |a_{t}^\text{lk} - a_{t-1}^\text{lk}|
\end{equation}
where $c_\text{exist}$ is a fixed existence penalty applied at every step, $p_{t}^{\text{ego}}$ denotes the longitudinal position of the ego vehicle, and $|a_{t}^\text{lk} - a_{t-1}^\text{lk}|$ represents the jerk. $\lambda_\text{prog}$ and $\lambda_\text{jerk}$ are weighting coefficients for progress and comfort, respectively.

\subsubsection{Safety and Gap Maintenance}
A critical component of the reward function is the continuous safety penalty, which enforces safe time headways with surrounding vehicles. If the gap to the leading or following vehicle falls below a safety threshold, a non-linear penalty is applied:
\begin{equation}
\begin{split}
    r_{\text{safe}} = &-\lambda_\text{gap} \sum_{k \in \{\text{ahead}, \text{behind}\}} \tanh\left(\frac{1}{|d_k| + \epsilon}\right) \cdot \\
    &\mathbb{I}(d_k < d_\text{safe})
\end{split}
\end{equation}
   
where $\lambda_\text{gap}$ scales the intensity of the safety penalty, $d_k$ is the distance to the neighbor $k$, $\epsilon$ is a small constant to prevent division by zero, and $\mathbb{I}(\cdot)$ is the indicator function. The hyperbolic tangent ensures the penalty saturates as the distance approaches zero.

\subsubsection{Event-Based and Terminal Rewards}
Sparse rewards are assigned for specific events, including successful merging, reaching the goal, or safety violations. The event reward $r_\text{event}$ is defined as:
\begin{equation}
\begin{aligned}
r_{\text{event}} =\;&
R_{\text{merge}} \mathbb{I}_{\text{merge}}
+ R_{\text{goal}} \mathbb{I}_{\text{goal}} \\
&- C_{\text{coll}} \mathbb{I}_{\text{coll}}
- C_{\text{lat}} \mathbb{I}_{\text{lat}}
- C_{\text{time}} \mathbb{I}_{\text{timeout}} .
\end{aligned}
\end{equation}
Here, $R_{\text{merge}}$ and $R_{\text{goal}}$ are positive terminal rewards for performing a merge and reaching the goal successfully, while
$C_{\text{coll}}$, $C_{\text{lat}}$, and $C_{\text{time}}$ penalize collisions,
unnecessary lane changes, and timeouts, respectively.

%% file: 04-Method/method.tex
We propose a unified control architecture that integrates the RDMDP state augmentation with a safety controller. An overview of the architecture is depicted in Figure~\ref{fig:framework}.
\begin{figure*}[t]
    \centering
    \includestandalone[width=0.9\textwidth]{Figures/network}
    \caption{{Overview of the DAROM framework.} To address random observation delays, the system augments the delayed observation $o_t$ with the masked action history $u_{t-\omega_t:t-1}$ and delay magnitude $\omega_t$. These inputs are fused into a latent representation $z_t$ by the \textit{Delay-Aware Encoder}. A unified soft actor-critic (SAC) agent then generates a raw control action $a_t$, which is validated and potentially overridden by the \textit{Safety Controller} ($a_{t_\text{safe}}$) to ensure collision-free merging.}
    \label{fig:framework}
\end{figure*}
\subsection{Augmented State Construction}
Consistent with Definition 1, our policy input is the augmented state $x_t = (o_t, u_t, \omega_t)$.
\begin{itemize}
    \item \textbf{Received Delayed Observation ($o_t$):} The kinematic state perceived by the RSU at time $t-\omega_t$. In our settings, the ego vehicle's state is instantaneous since we assume it can rely on the onboard sensors to calculate its real-time position and velocity.  
    \item \textbf{Action Buffer ($u_t$):} A history of actions $\{a_{t-k}, a_{t-k+1}, \dots, a_{t-1}\}$. To strictly assist the network in causal inference, we apply a mask to this buffer. Actions older than the current delay (i.e., $a_{t-k}$ where $k > \omega_t$) are masked to zero. This explicitly indicates to the network which actions in the buffer have effectively contributed to the evolution from $o_t$ to the current time.
    \item \textbf{Delay Indicator ($\omega_t$):} The scalar value of the received observation delay magnitude.
\end{itemize}
To better illustrate the augmented state construction, we included an example in Table \ref{tab:example} which represents an 8-time-step scenario. Figure \ref{fig:delay} further elaborates on time-step $t=6$. Here, $d_t$ represents the delay of state $s_t$ and $s_\text{hist}$ represents the unobserved states, where each state ${s_{t}}^{t + d_t}$ will be observed at time $t + d_t$.
\setlength{\tabcolsep}{3pt}
\renewcommand{\arraystretch}{1.25}
\begin{table}[h]
\caption{An augmented state construction example for a scenario with 8 time-steps. For clarity, we truncate the action buffer and omit zeros. Figure \ref{fig:delay} further elaborates on time-step~$t = 6$.}
\label{tab:example}
\resizebox{\linewidth}{!}{%
\rowcolors{3}{stripe}{white}
\centering

\begin{tabular}{c|cccccccc}
\midrule
\rowcolor{headerblue}
\color{white}\textbf{$t$} &
\color{white}\textbf{1} & \color{white}\textbf{2} & \color{white}\textbf{3} & \color{white}\textbf{4} &
\color{white}\textbf{5} & \color{white}\textbf{6} & \color{white}\textbf{7} & \color{white}\textbf{8} \\
\midrule

$d_t$ &
$0$ & $1$ & $0$ & $5$ & $4$ & $2$ & $1$ & $0$ \\

$s_{\text{hist}}$ &
$\big[{s_{1}^{1}}\big]$ &
$\big[s_{2}^{3}\big]$ &
$\big[{s_{3}^{3}}\big]$ &
$\big[s_{4}^{9}\big]$ &
$\big[s_{5}^{9},s_{4}^{9}\big]$ &
$\big[s_{5}^{9},s_{6}^{8},s_{4}^{9}\big]$ &
$\big[s_{5}^{9},s_{6}^{8},{s_{4}^{9}}\big]$ &
$\big[s_{5}^{9},{s_{8}^{8}}\big]$ \\

$o_t$ &
$s_{1}$ & $s_{1}$ & $s_{3}$ & $s_{3}$ & $s_{3}$ & $s_{3}$ & $s_{4}$ & $s_{8}$ \\

$u_t$ &
$\big[\big]$ &
$\big[a_{1}\big]$ &
$\big[\big]$ &
$\big[a_{3}\big]$ &
$\big[a_{3},a_{4}\big]$ &
$\big[a_{3},a_{4},a_{5}\big]$ &
$\big[a_{4},a_{5},a_{6}\big]$ &
$\big[\big]$ \\

$\omega_t$ & 0 & 1 & 0 & 1 & 2 & 3 & 3 & 0 \\

\bottomrule
\end{tabular}
}
\end{table}
\begin{figure}[!h]
    \centering
    \includestandalone[width=0.7\columnwidth]{Figures/environment}

    \caption{{Augmented state construction details.} At time-step $t=6$, the state $s_6$ has a delay of $d_6 = 2$, and the received observation $o_6 = s_3$ has a delay of $\omega_6 = 3$. The ego vehicle still has not observed $s_4$ and $s_5$ because of their delay amounts. $x_6$ represents the augmented state which includes delayed observation, action buffer, and the delay magnitude for the received observation.}
    \label{fig:delay}
\end{figure}

\subsection{Delay-Aware Encoder Network}
To process the heterogeneous structure of the delayed input $x_t$, we introduce a \textit{Delay-Aware Encoder} that maps delayed observations, action history, and delay information into a compact latent representation suitable for policy learning. At each decision step, the encoder receives a concatenated input
$
x_t = [o_t, u_t, \omega_t].
$

The encoder first separates these components and processes them through specialized representation streams. The delayed observation $o_t$ is mapped to a fixed-dimensional embedding through a multilayer perceptron (MLP), capturing the instantaneous physical state information. The action history $u_t$ is encoded using one of several architectural choices to capture different levels of temporal structure, while the delay $\omega_t$ is provided explicitly to condition the latent representation on the magnitude of observation staleness. The resulting embeddings are fused through a shared MLP to produce the latent state representation used by the policy and value networks.

In the \textbf{MLP-based encoder}, the action history is treated as a flattened vector and concatenated directly with the observation embedding and delay scalar. This design provides a lightweight representation that ignores temporal dependencies within the action buffer.

In the \textbf{GRU-based encoder}, the action buffer is reshaped into a sequence and processed by a gated recurrent unit (GRU), with the final hidden state serving as a summary of past control actions. This enables the encoder to model temporal dependencies induced by delayed execution and observation. The GRU output is then fused with the observation embedding and delay information.

In the \textbf{Transformer-based encoder}, the observation stream consists of a sequence of entity-level features, which are processed using self-attention to capture interactions among surrounding vehicles. The action history and delay are encoded separately using MLPs and fused with the Transformer-derived observation representation. This architecture emphasizes relational reasoning among traffic participants while conditioning on past actions and delay magnitude.

Across all variants, explicitly conditioning the encoder on $\omega_t$ allows the policy to adapt its behavior based on the severity of communication delay, thereby mitigating the violation of the Markov assumption caused by delayed and partially observable state information.

\subsection{Physics-Based Safety Controller}
To ensure robust safety under uncertainty, we implement a safety controller that intercepts the agent's action $a_t$ before execution.
Let $v_{rel}$ be the relative velocity between the ego vehicle and a conflicting vehicle. The required stopping distance is calculated as:
\begin{equation}
    d_\text{stop} = \frac{\max(0, -v_{rel})^2}{2 |a_\text{brake}|}
\end{equation}
If the current gap $g < d_\text{stop} + g_\text{min}$, the proposed action $a_t$ is overridden by a safe fallback $a_{t_\text{safe}}$ (e.g., maximum braking or lane keeping). Here $g_\text{min}$ denotes the minimum desired gap.

\subsection{Learning Algorithm}
We employ SAC to learn the optimal policy over the augmented space $\mathcal{X}$. The objective is to maximize the entropy-regularized return:
\begin{equation}
    J(\pi) = \sum_{t=0}^{T} \mathbb{E}_{(x_t, a_t) \sim \rho_\pi} [\gamma^t  (r(x_t, a_t) + \alpha \mathcal{H}(\pi(\cdot|x_t)))]
\end{equation}
Here, $\rho_{\pi}$ denotes the state--action occupancy measure induced by policy $\pi$, $\mathcal{H}(\pi(\cdot|x_t))$ is the entropy of the action distribution at augmented state $x_t$, and $\alpha$ is the temperature coefficient that trades off reward maximization and entropy regularization.
By training on the augmented state $x_t$, the agent learns to implicitly perform belief-state updates, correlating the delayed observation $o_t$, the action history $u_t$, and the delay magnitude to approximate the optimal action for the unobserved true state.

%% file: Figures/environment.tex
\tikzset{every picture/.style={line width=0.75pt}} 

\begin{tikzpicture}[x=0.75pt,y=0.75pt,yscale=-1,xscale=1]

\draw  [line width=2.25]  (24,21.5) .. controls (24,17.08) and (27.58,13.5) .. (32,13.5) -- (78,13.5) .. controls (82.42,13.5) and (86,17.08) .. (86,21.5) -- (86,221.5) .. controls (86,225.92) and (82.42,229.5) .. (78,229.5) -- (32,229.5) .. controls (27.58,229.5) and (24,225.92) .. (24,221.5) -- cycle ;
\draw  [fill={rgb, 255:red, 255; green, 227; blue, 189 }  ,fill opacity=1 ] (31,149) -- (77,149) -- (77,174.5) -- (31,174.5) -- cycle ;
\draw  [fill={rgb, 255:red, 221; green, 231; blue, 255 }  ,fill opacity=1 ] (55.25,179.5) -- (78,202.25) -- (55.25,225) -- (32.5,202.25) -- cycle ;
\draw  [fill={rgb, 255:red, 255; green, 227; blue, 189 }  ,fill opacity=1 ] (31,123.5) -- (77,123.5) -- (77,149) -- (31,149) -- cycle ;
\draw  [fill={rgb, 255:red, 255; green, 227; blue, 189 }  ,fill opacity=1 ] (31,98) -- (77,98) -- (77,123.5) -- (31,123.5) -- cycle ;
\draw  [fill={rgb, 255:red, 255; green, 227; blue, 189 }  ,fill opacity=1 ] (31,72.5) -- (77,72.5) -- (77,98) -- (31,98) -- cycle ;
\draw  [fill={rgb, 255:red, 255; green, 227; blue, 189 }  ,fill opacity=1 ] (31,47) -- (77,47) -- (77,72.5) -- (31,72.5) -- cycle ;
\draw   (31.2,24.3) .. controls (31.2,20.99) and (33.89,18.3) .. (37.2,18.3) -- (71.2,18.3) .. controls (74.51,18.3) and (77.2,20.99) .. (77.2,24.3) -- (77.2,37.8) .. controls (77.2,41.11) and (74.51,43.8) .. (71.2,43.8) -- (37.2,43.8) .. controls (33.89,43.8) and (31.2,41.11) .. (31.2,37.8) -- cycle ;
\draw  [fill={rgb, 255:red, 221; green, 231; blue, 255 }  ,fill opacity=1 ] (53.65,243.9) -- (76.4,266.65) -- (53.65,289.4) -- (30.9,266.65) -- cycle ;
\draw  [fill={rgb, 255:red, 221; green, 231; blue, 255 }  ,fill opacity=1 ] (91.65,298) -- (114.4,320.75) -- (91.65,343.5) -- (68.9,320.75) -- cycle ;
\draw (205.7,224.51) node  {\includegraphics[width=144.91pt,height=102.44pt]{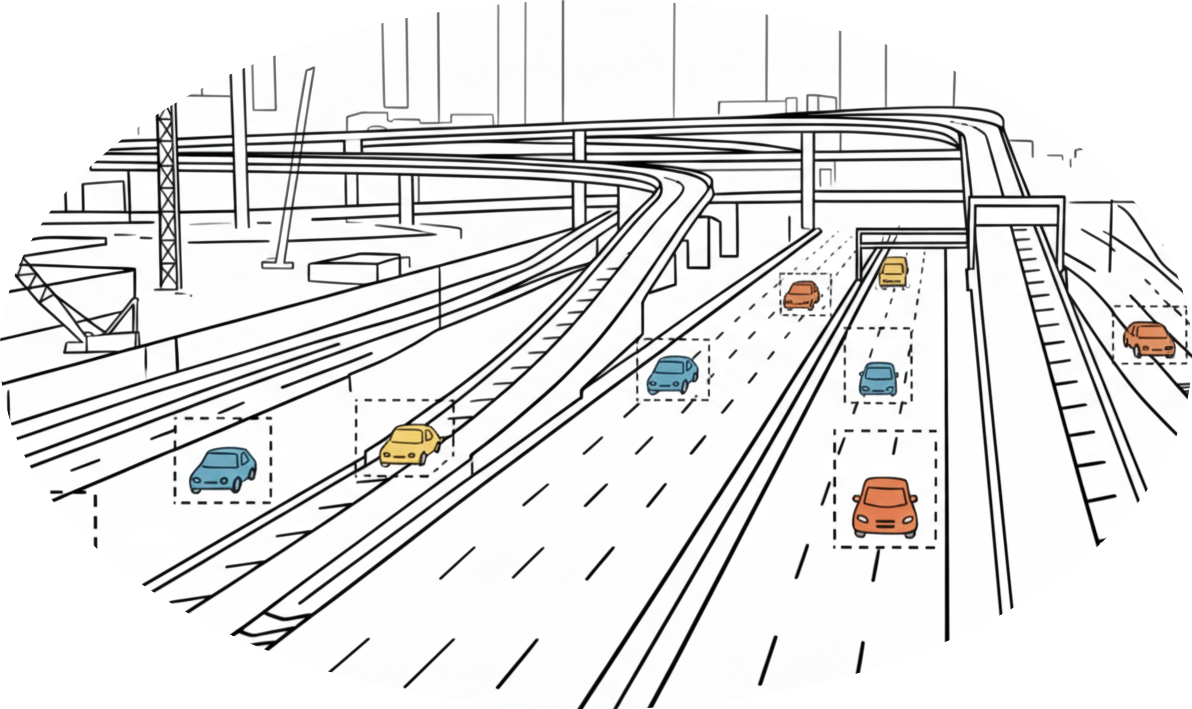}};
\draw  [line width=1.5]  (108.7,223.85) .. controls (108.7,186.01) and (152.58,155.33) .. (206.7,155.33) .. controls (260.82,155.33) and (304.7,186.01) .. (304.7,223.85) .. controls (304.7,261.68) and (260.82,292.36) .. (206.7,292.36) .. controls (152.58,292.36) and (108.7,261.68) .. (108.7,223.85) -- cycle ;

\draw  [color={rgb, 255:red, 164; green, 158; blue, 133 }  ,draw opacity=1 ][fill={rgb, 255:red, 255; green, 79; blue, 79 }  ,fill opacity=1 ][line width=0.75]  (176.96,17.8) .. controls (176.95,14.69) and (179.46,12.16) .. (182.57,12.15) -- (226.14,12.02) .. controls (229.25,12.01) and (231.78,14.52) .. (231.79,17.63) -- (231.85,34.53) .. controls (231.86,37.64) and (229.34,40.17) .. (226.23,40.18) -- (182.67,40.32) .. controls (179.56,40.33) and (177.03,37.81) .. (177.02,34.7) -- cycle ;
\draw  [color={rgb, 255:red, 133; green, 153; blue, 164 }  ,draw opacity=1 ][fill={rgb, 255:red, 255; green, 255; blue, 255 }  ,fill opacity=1 ][line width=0.75]  (214.79,16.72) -- (214.68,16.72) -- (214.79,16.67) -- (214.79,16.72) -- cycle (201.54,16.76) -- (201.53,13.29) -- (213.95,13.25) -- (206.81,16.75) -- (201.54,16.76) -- cycle ;
\draw  [color={rgb, 255:red, 133; green, 153; blue, 164 }  ,draw opacity=1 ][fill={rgb, 255:red, 255; green, 255; blue, 255 }  ,fill opacity=1 ][line width=0.75]  (214.71,35.61) -- (214.59,35.61) -- (214.71,35.67) -- (214.71,35.61) -- cycle (201.46,35.65) -- (201.47,39.13) -- (213.89,39.09) -- (206.73,35.64) -- (201.46,35.65) -- cycle ;
\draw  [color={rgb, 255:red, 133; green, 153; blue, 164 }  ,draw opacity=1 ][fill={rgb, 255:red, 255; green, 255; blue, 255 }  ,fill opacity=1 ][line width=0.75]  (199.54,13.29) -- (199.55,16.92) -- (190.29,16.95) -- (187.81,13.33) -- (199.54,13.29) -- cycle ;
\draw  [color={rgb, 255:red, 133; green, 153; blue, 164 }  ,draw opacity=1 ][fill={rgb, 255:red, 255; green, 255; blue, 255 }  ,fill opacity=1 ][line width=0.75]  (199.48,39.29) -- (199.47,35.66) -- (190.21,35.69) -- (187.75,39.32) -- (199.48,39.29) -- cycle ;
\draw  [color={rgb, 255:red, 133; green, 153; blue, 164 }  ,draw opacity=1 ][fill={rgb, 255:red, 255; green, 255; blue, 255 }  ,fill opacity=1 ][line width=0.75]  (209.22,34.59) -- (209.17,17.79) -- (219.39,14.89) -- (219.52,14.89) .. controls (220.5,17.81) and (221.08,21.35) .. (221.1,25.17) .. controls (221.11,29.81) and (220.28,34.05) .. (218.9,37.26) -- (209.22,34.59) -- cycle ;
\draw  [color={rgb, 255:red, 133; green, 153; blue, 164 }  ,draw opacity=1 ][fill={rgb, 255:red, 255; green, 255; blue, 255 }  ,fill opacity=1 ][line width=0.75]  (184.44,15) -- (188.44,17.98) -- (188.5,34.66) -- (184.52,37.67) -- cycle ;

\draw  [fill={rgb, 255:red, 221; green, 231; blue, 255 }  ,fill opacity=1 ] (139.35,246.3) -- (162.1,269.05) -- (139.35,291.8) -- (116.6,269.05) -- cycle ;
\draw  [draw opacity=0][line width=3.75]  (180.28,125.86) .. controls (181.57,115.43) and (191.41,107.32) .. (203.35,107.32) .. controls (215.2,107.32) and (224.97,115.29) .. (226.39,125.59) -- (203.35,128.19) -- cycle ; \draw  [color={rgb, 255:red, 255; green, 206; blue, 118 }  ,draw opacity=1 ][line width=3.75]  (180.28,125.86) .. controls (181.57,115.43) and (191.41,107.32) .. (203.35,107.32) .. controls (215.2,107.32) and (224.97,115.29) .. (226.39,125.59) ;  
\draw  [draw opacity=0][line width=3.75]  (171.83,102.42) .. controls (174.59,89.04) and (187.45,78.92) .. (202.91,78.92) .. controls (218.14,78.92) and (230.86,88.75) .. (233.86,101.85) -- (202.91,107.79) -- cycle ; \draw  [color={rgb, 255:red, 255; green, 218; blue, 153 }  ,draw opacity=1 ][line width=3.75]  (171.83,102.42) .. controls (174.59,89.04) and (187.45,78.92) .. (202.91,78.92) .. controls (218.14,78.92) and (230.86,88.75) .. (233.86,101.85) ;  
\draw  [draw opacity=0][line width=3.75]  (163.21,79.89) .. controls (168.48,63.98) and (184.27,52.43) .. (202.93,52.43) .. controls (221.21,52.43) and (236.74,63.51) .. (242.32,78.91) -- (202.93,91.52) -- cycle ; \draw  [color={rgb, 255:red, 255; green, 231; blue, 193 }  ,draw opacity=1 ][line width=3.75]  (163.21,79.89) .. controls (168.48,63.98) and (184.27,52.43) .. (202.93,52.43) .. controls (221.21,52.43) and (236.74,63.51) .. (242.32,78.91) ;  
\draw  [draw opacity=0][line width=3.75]  (189.56,145.31) .. controls (190.36,138.82) and (196.47,133.78) .. (203.9,133.78) .. controls (211.26,133.78) and (217.34,138.74) .. (218.22,145.14) -- (203.9,146.76) -- cycle ; \draw  [color={rgb, 255:red, 255; green, 192; blue, 77 }  ,draw opacity=1 ][line width=3.75]  (189.56,145.31) .. controls (190.36,138.82) and (196.47,133.78) .. (203.9,133.78) .. controls (211.26,133.78) and (217.34,138.74) .. (218.22,145.14) ;  

\draw (45,194.4) node [anchor=north west][inner sep=0.75pt]    {\Large$s_{3}$};
\draw (43.4,259.8) node [anchor=north west][inner sep=0.75pt]    {\Large$s_{4}$};
\draw (44.4,159.4) node [anchor=north west][inner sep=0.75pt]    {$\dots $};
\draw (48.4,130.4) node [anchor=north west][inner sep=0.75pt]    {\Large$0$};
\draw (46.4,106.4) node [anchor=north west][inner sep=0.75pt]    {\Large$a_{5}$};
\draw (45.4,79.4) node [anchor=north west][inner sep=0.75pt]    {\Large$a_{4}$};
\draw (45.4,55.4) node [anchor=north west][inner sep=0.75pt]    {\Large$a_{3}$};
\draw (34,24.9) node [anchor=north west][inner sep=0.75pt]  [font=\normalsize]  {$\omega _{6} =3$};
\draw (27.4,288.8) node [anchor=north west][inner sep=0.75pt]    {\Large$d_{4} =5$};
\draw (81.4,313.9) node [anchor=north west][inner sep=0.75pt]    {\Large$s_{5}$};
\draw (63.4,343.4) node [anchor=north west][inner sep=0.75pt]    {\Large$d_{5} =4$};
\draw (129.6,263.2) node [anchor=north west][inner sep=0.75pt]    {\Large$s_{6}$};
\draw (115.2,302.3) node [anchor=north west][inner sep=0.75pt]   [align=left] {\Large Undelayed Environment};
\draw (248.83,95.27) node [anchor=north west][inner sep=0.75pt]    {\Large$d_{6} =2$};
\draw (189.33,-6.67) node [anchor=north west][inner sep=0.75pt]   [align=left] {\large ego};
\draw (111.1,159.62) node [anchor=north west][inner sep=0.75pt]    {\Large $t_{6}$};
\draw (45.4,-5) node [anchor=north west][inner sep=0.75pt]    {\Large$x_{6}$};

\end{tikzpicture}

%% file: 05-Experiments/experiments.tex
\subsection{Experimental Setup}
All simulations are performed in simulation of urban mobility (SUMO). To ensure the algorithm is capable of tackling real-world challenges, we extracted traffic demand (i.e., lane-level demand) and the surrounding vehicles' desired maximum speeds from the NGSIM US Highway 101 dataset \cite{ngsim}, representing mainstream highway conditions.

In all experiments, the state representation supports a maximum of 30 surrounding vehicles. The surrounding vehicles are sorted based on their relative lane position and longitudinal distance to the ego vehicle (see Section~\ref{sec:formulation:state_action}). This sorting mechanism ensures that, even when the traffic density exceeds the buffer size, the agent always captures the closest and most critical neighbors required for safe maneuvering.

To simulate heterogeneous driving behaviors, we defined two distinct driver profiles: aggressive and cooperative. The parameters for these vehicles are sampled as follows:
\begin{itemize}
    \item \textbf{Aggressive Drivers:} These drivers represent risky behavior. Their minimum time headway ($\tau$) is sampled from a uniform distribution $\mathcal{U}(0.1, 0.7)$ s, and their maximum speed is sampled from $\mathcal{U}(10, 13)$ m/s. To strictly enforce a risky driving style, all safety regulations for aggressive drivers are turned off in the SUMO setting.
    \item \textbf{Cooperative Drivers:} These drivers represent conservative traffic. Their minimum time headway is sampled from $\mathcal{U}(0.6, 0.8)$ s, with maximum speeds sampled from $\mathcal{U}(8, 11)$ m/s.
    \item \textbf{Ego Vehicle:} To ensure that our proposed algorithm is completely in charge of controlling the vehicle's actions, the safety regulations for the ego vehicle are also suppressed.
\end{itemize}
\begin{table}[h]
\centering
\begin{threeparttable}
\captionsetup{width=\columnwidth,justification=raggedright,singlelinecheck=false}
\caption{Traffic flow rates per lane for different difficulty modes.}
\label{tab:traffic_flow}
\begin{tabular}{@{}cccc@{}}
\toprule
\textbf{Lane} & \textbf{Easy (vph)} & \textbf{Medium (vph)} & \textbf{Hard (vph)} \\ \midrule
1 & 360 & 720 & 1394 \\
2 & 360 & 684 & 1460 \\
3 & 360 & 684 & 1390 \\
4 & 360 & 684 & 1374 \\
5 & 360 & 684 & 1490 \\ \bottomrule
\end{tabular}
\begin{tablenotes}[flushleft]
\footnotesize
\item vph denotes vehicles per hour.
\end{tablenotes}
\end{threeparttable}
\end{table}


\textbf{Training Details:} All agents were trained for 7,000,000 time-steps with a learning rate of $3 \times 10^{-5}$ and a batch size of $512$. Apart from the mentioned hyperparameters, we use Stable-Baselines3's default SAC setting for the actor-critic network architectures (2 hidden layers with 256 units) \cite{stable-baselines3}. The training process is optimized using the Adam optimizer and executed on a high-performance workstation equipped with an AMD 16-Core CPU and an NVIDIA RTX 5090 GPU to accelerate computation.

\subsection{Network Architectures and Baselines}
To investigate the optimal method for processing the augmented state $x_t = (o_t, u_t, \omega_t)$, we implemented and compared three variants of the Delay-Aware Encoder, extracting parameters directly from our implementation:

\begin{itemize}
    \item \textbf{MLP:} In this variant, the observation encoder consists of two linear layers with 64 units, LayerNorm, and ReLU activations. The fusion network contains two linear layers (128 and 256 units) with LayerNorm and Tanh activations. 
    
    \item \textbf{GRU:} Utilizing a GRU to process the action history $u_t$. The GRU has a hidden state size of 64. The observation encoder uses two linear layers with 64 units, LayerNorm, and ReLU activations. The fusion network contains two linear layers (128 and 256 units) with LayerNorm and Tanh activations. 
    
    \item \textbf{Transformer:} Employing a Transformer-based architecture. The entity encoder utilizes a transformer with 1 attention head, and a feedforward dimension of 64 across 1 transformer layer with a model dimension of 32. The action history and delay are encoded through separate linear layers projecting to 32 dimensions. The output projection maps to 128-dimensional features. 
     
\end{itemize}

\textbf{Baselines:} We compare the proposed DAROM architectures against the following methods:
\begin{itemize}
    \item \textbf{No Encoder:} An ablation baseline where the agent processes the raw $x_t$ without any encoder. It excludes the separate processing of the received observation, isolating the impact of the Delay-Aware Encoder on performance.
    
    \item \textbf{MPC:} The MPC baseline is implemented as a three-phase controller for highway on-ramp merging. The current phase is determined from the ego longitudinal position and a merge flag: phase 1 (pre-merge), phase 2 (merge zone), and phase 3 (post-merge). At each step, ego motion is predicted with a discrete-time kinematic model driven by an acceleration sequence over a fixed horizon, while surrounding vehicles are predicted under constant velocity. Before planning, each neighbor's delayed position is corrected as $\hat{x}^i_{\text{rel}} = x^i_{\text{rel}} + v^i_{\text{rel}} \cdot \omega_t \cdot \Delta t$, giving the MPC a current-state estimate without modifying the ego state, which is always available from onboard sensors. The objective in phases 1 and 3 is quadratic in speed error (reference speed tracking) and acceleration (smoothness), plus a large penalty for longitudinal gaps below a minimum safe distance. In phase 2, the same terms apply and the cost additionally strongly encourages merging; the controller enumerates a small set of candidate lane-change sequences (no change or a single merge-right at some step). For each lane sequence, a bounded nonlinear program over the acceleration sequence is solved with L-BFGS-B; the acceleration and lane-change sequences with lowest cost are chosen and only their first step is applied, yielding a receding-horizon acceleration and discrete lane-change command.
\item \textbf{DRL-ORMOC \cite{9922428}:} A state-of-the-art deep reinforcement learning baseline for highway on-ramp merging. The method operates on the most recent observation and does not incorporate state augmentation or explicit mechanisms to compensate for observation delays. \textit{Since the original codebase is not publicly available, we re-implemented the algorithm within our simulation framework and tuned the reward hyperparameters to achieve its best performance. For consistency across methods, we replaced the original safety controller with our physics-based safety module. {Due to differences in reward scale, final return values were mapped to match our framework's reward range}; traffic-level performance metrics are additionally reported to enable a fair and meaningful comparison.}

\end{itemize}

\subsection{Delay Profiles and Stochastic Latency Modeling}

We simulate the RSU processing delay as a random variable $\omega_t \in [0, \Omega_{\max}]$.
In all experiments, $\Omega_{\max} = 20$ time-steps, which at $\Delta t = 0.1s$\, corresponds to
a maximum latency of 2.0 seconds. While typical RSU perception-to-vehicle delivery under nominal conditions is within    
  $\approx$100 ms ~\cite{s20185320}, we set $\Omega_\text{max} = 2.0s$ as a 
  deliberate worst-case stress test, capturing edge-computing overload, packet             
  retransmission under congestion, and multi-hop relaying scenarios; the agent is evaluated against this
  conservative envelope to demonstrate robustness beyond typical operating conditions. We trained DAROM-GRU only on the uniform delay distribution and then evaluated it across five stochastic delay distributions to assess robustness and cross-distribution generalization under different V2I communication regimes:

\begin{itemize}
  \item \textbf{Uniform:} $\omega_t \sim \mathcal{U}(0, \Omega_{\max})$, the baseline training
        distribution providing equal probability mass across all latencies.

  \item \textbf{Bimodal:} The delay is sampled from a two-mode discrete mixture representing good and bad connection regimes. With probability $0.6$, we sample
        $\omega_t \sim \mathcal{U}_{\mathrm{}}\bigl(0,\, \max(\lfloor \Omega_{\max}/5 \rfloor, 1)\bigr)$,
        corresponding to low-delay operation. With probability $0.4$, we sample
        $\omega_t \sim \mathcal{U}_{\mathrm{}}\bigl(\lfloor 3\Omega_{\max}/5 \rfloor,\, \Omega_{\max}\bigr)$,
        corresponding to high-delay operation. This models a network that alternates between stable low-latency periods and degraded high-latency conditions.

\item \textbf{Bursty:} The delay is governed by a Markov regime-switching process with a
        hidden binary congestion state $c_t \in \{0, 1\}$ and transition probabilities
        $P(c_t{=}1 \mid c_{t-1}{=}1) = 0.9$ and $P(c_t{=}1 \mid c_{t-1}{=}0)
         = 0.05$.
        The delay is then drawn conditionally as
        \begin{equation*}
          \omega_t \sim
          \begin{cases}
            \mathcal{U}(0,\; \lfloor \Omega_{\max}/4 \rfloor) & \text{if } c_t = 0 \text{ (normal)},\\
            \mathcal{U}(\lfloor \Omega_{\max}/2 \rfloor,\; \Omega_{\max}) & \text{if } c_t = 1 \text{ (congested)}.
          \end{cases}
        \end{equation*}
        This models persistent congestion episodes with rare but sustained high-latency bursts.

  \item \textbf{Triangular:} $\omega_t \sim \mathrm{Triangular}(0,\, \lfloor \Omega_{\max}/2 \rfloor,\, \Omega_{\max})$,
        parameterized by lower bound, mode, and upper bound respectively. This symmetric
        unimodal distribution concentrates mass around moderate latencies, modeling a network
        operating near its design capacity.

  \item \textbf{Exponential:} $\omega_t \sim \mathrm{Exp}(\beta) \mid \omega_t \leq \Omega_{\max}$,
        i.e., an exponential distribution with rate $\beta = \lfloor \Omega_{\max}/3 \rfloor$ truncated to $[0, \Omega_{\max}]$,
        placing high probability on low latencies with a decaying tail, reflecting typical queuing
        behavior in lightly loaded networks.
\end{itemize}

\subsection{Ablation Study: Augmented State Components and Delay Compensation}
To understand the contribution of specific components within the augmented state, we conducted an ablation study specifically in the Hard scenario. We compared the full DAROM-GRU model against four partial configurations:
\begin{enumerate}
    \item \textbf{Delayed State Only:} The agent only observes $o_t$, lacking both the action history $u_t$ and the received observation delay magnitude $\omega_t$.
    \item \textbf{Delayed State with Delay Magnitude:} The agent observes $o_t$ and  $\omega_t$, but lacks the action history $u_t$.
    \item \textbf{Delayed State with Action Buffer:} The agent observes $o_t$ and $u_t$, but lacks the delay magnitude $\omega_t$.
    \item \textbf{Linear State Predictor:} A constant-velocity baseline that predicts the current position of each surrounding vehicle as $\hat{x}^i_{\text{rel}} = x^i_{\text{rel}} + v^i_{\text{rel}} \cdot \omega_t \cdot \Delta t$, using the received delay magnitude $\omega_t$ and holding lateral position and velocity constant. The policy receives only the predicted state, without access to the action buffer or delay indicator.
\end{enumerate}

\subsection{Results and Discussion}

\begin{figure*}[!h]
    \centering

    \includegraphics[width=0.8\linewidth]{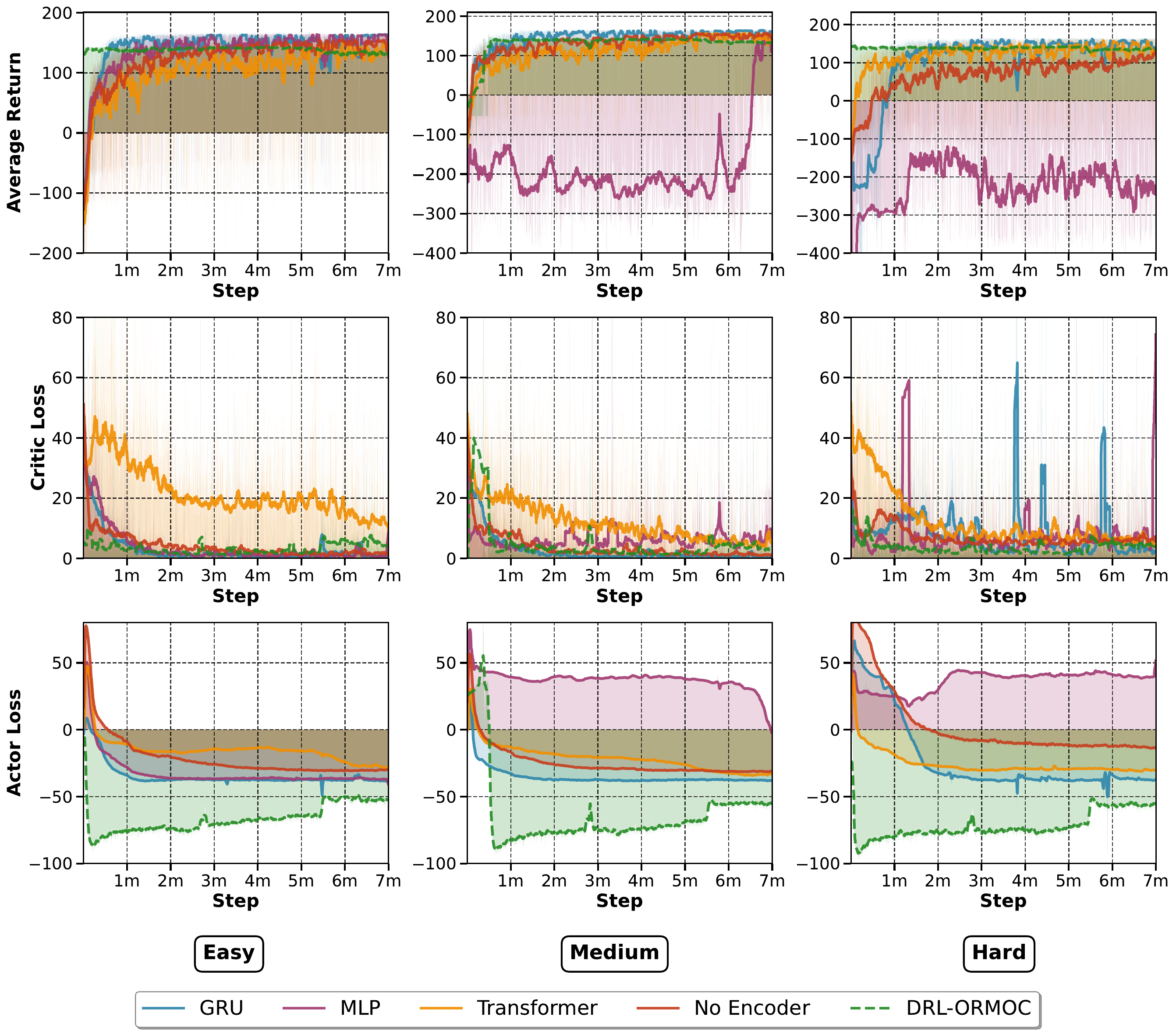}

    \caption{Overall comparison across Easy, Medium, and Hard modes. The GRU encoder captures temporal dependencies in the action history through recurrent hidden states, the MLP encoder treats the action buffer as a flat vector and ignores temporal order, and the Transformer encoder uses self-attention over entity tokens to model inter-vehicle interactions. Rows correspond to average return, critic loss, and actor loss, respectively. The horizontal axis denotes the number of environment interaction steps in millions ($1\text{m} = 10^6$)}
    \label{fig:main_performance}
\end{figure*}


\begin{figure}[!h]
    \centering
    \includegraphics[width=\columnwidth]{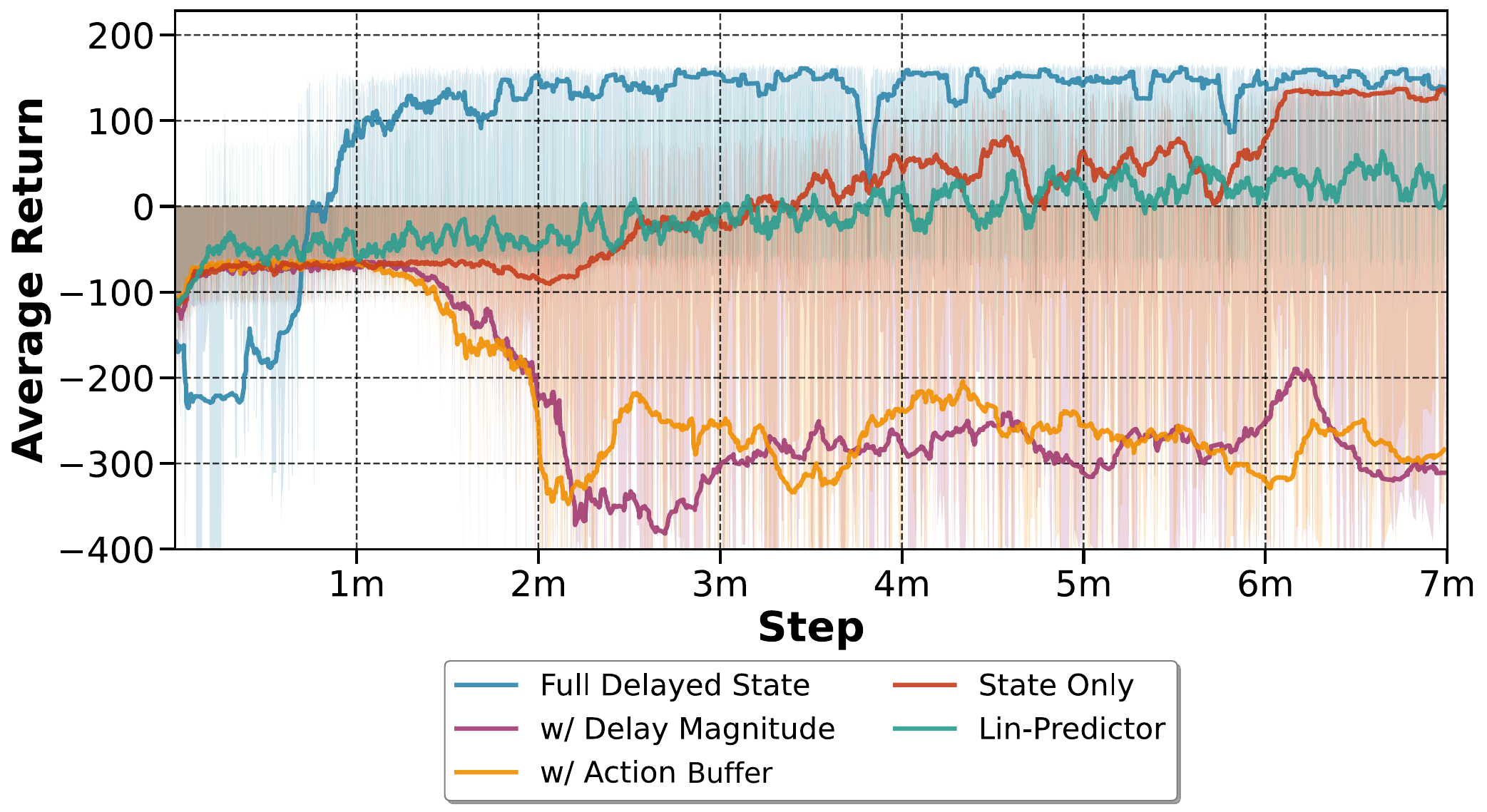}
    \caption{Ablation study in Hard mode. Comparing the full DAROM-GRU architecture against ablated inputs, namely delayed state, delayed state with delay magnitude, delayed state with action buffer, and the Linear State Predictor baseline. Two ablated variants fail to converge, the Linear State Predictor converges to a suboptimal policy, and the full model achieves the best performance. The horizontal axis denotes the number of environment interaction steps in millions ($1\text{m} = 10^6$).}
    \label{fig:ablation_hard}
\end{figure}
Figure \ref{fig:main_performance} summarizes the training dynamics for different network architectures and baselines, while Figure \ref{fig:ablation_hard} reports the feature-importance ablation and delay compensation strategy results. In Figure \ref{fig:main_performance}, DAROM-GRU exhibits a steady increase in average return followed by a stable plateau, while critic and actor losses decrease and stabilize, indicating convergence. In contrast, weaker baselines show noisier learning curves and less consistent improvements. Figure \ref{fig:ablation_hard} further highlights the importance of the full augmented state: the full model reaches and maintains higher returns, whereas ablated variants either stagnate at low returns or fail to converge.

We evaluated the trained policies over 500 evaluation episodes, repeated across three random seeds. The comparative performance metrics are detailed in Table \ref{tab:main_results} for different network architectures and baselines. The ablation study results are presented in Table \ref{tab:ablation_results} and the evaluations for different delay distributions are reported in Table \ref{tab:delay_profile_results}.

\begin{table*}[!h]
\centering
\begin{threeparttable}
\caption{Comparative performance analysis of DAROM framework variants and baselines across Easy, Medium, and Hard traffic scenarios.}
\label{tab:main_results}

\setlength{\tabcolsep}{3pt} 
\begin{tabular}{@{}llcccccc@{}}

\toprule
\textbf{Difficulty} & \textbf{Method} & \textbf{Success Rate (\%)} & \textbf{Collision Rate (\%)} & \textbf{No Merge Rate (\%)} & \textbf{Avg. Return} & \textbf{Avg. Ego Vel.} & \textbf{Avg. Jerk} \\ \midrule
\multirow{7}{*}{\textbf{Easy}} 
 & MPC & $83.37 \pm 3.07$ & $16.30 \pm 3.40$ & $0.33 \pm 0.35$ &  - & $19.25 \pm 0.04$ & $-0.39 \pm 0.04$ \\
 & DRL-ORMOC & $96.61 \pm 0.90$ & $3.39 \pm 0.90$ & \bm{$0.00 \pm 0.00$} & $143.58 \pm 0.58$ & $27.94 \pm 0.08$ & $-0.18 \pm 0.01$ \\
 & No Encoder & $98.00 \pm 0.39$ & $0.53 \pm 0.26$ & $1.46 \pm 0.35$ & $155.81 \pm 1.23$ & $28.82 \pm 0.06$ & {$-0.23 \pm 0.01$} \\
 & \textbf{DAROM-MLP}& \bm{$99.60 \pm 0.00$} & \bm{$0.20 \pm 0.00$} & $0.20 \pm 0.00$ & $172.11 \pm 0.57$ & \bm{$29.42 \pm 0.02$} & $-0.30 \pm 0.01$ \\
 & DAROM-GRU w/o SC & $91.68 \pm 1.36$ & $8.18 \pm 1.48$ & $0.13 \pm 0.13$ & $159.16 \pm 3.38$ & $28.82 \pm 0.13$ & \bm{$-0.08 \pm 0.01$} \\
 & DAROM-GRU & $98.74 \pm 0.26$ & $0.40 \pm 0.23$ & $0.13 \pm 0.26$ & \bm{$173.94 \pm 0.61$} & $29.31 \pm 0.06$ & $-0.27 \pm 0.01$ \\
 & DAROM-Transformer & $82.83 \pm 0.98$ & $0.40 \pm 0.45$ & $16.77 \pm 1.37$ & $127.50 \pm 2.79$ & $28.78 \pm 0.05$ & $-0.25 \pm 0.01$ \\ \midrule
\multirow{7}{*}{\textbf{Medium}} 
 & MPC & $83.63 \pm 0.81$ & $16.30 \pm 0.86$ & $0.07 \pm 0.13$ &  - & $19.23 \pm 0.04$ & $-0.35 \pm 0.02$ \\
 & DRL-ORMOC & $94.88 \pm 1.28$ & $4.99 \pm 1.04$ & $0.13 \pm 0.26$ & $142.03 \pm 0.74$ & $27.83 \pm 0.09$ & \bm{$-0.19 \pm 0.01$} \\
 & No Encoder & $99.07 \pm 0.52$ & $0.13 \pm 0.26$ & $0.80 \pm 0.45$ & $164.57 \pm 0.83$ & $29.37 \pm 0.03$ & $-0.26 \pm 0.01$ \\
 & DAROM-MLP & $69.59 \pm 1.24$ & $0.60 \pm 0.23$ & $29.81 \pm 1.16$ & $96.05 \pm 2.61$ & $27.79 \pm 0.10$ & $-0.25 \pm 0.01$ \\
 & DAROM-GRU w/o SC & $85.16 \pm 1.63$ & $14.84 \pm 1.63$ & \bm{$0.00 \pm 0.00$} & $142.15 \pm 4.46$ & $28.19 \pm 0.17$ & $-0.25 \pm 0.02$ \\
 & \textbf{DAROM-GRU} & \bm{$99.87 \pm 0.13$} & \bm{$0.00 \pm 0.00$} & $0.13 \pm 0.13$ & \bm{$178.57 \pm 0.32$} & \bm{$29.57 \pm 0.00$} & $-0.36 \pm 0.01$ \\
 & DAROM-Transformer & $84.36 \pm 1.84$ & $0.20 \pm 0.00$ & $15.44 \pm 1.84$ & $134.72 \pm 4.64$ & $28.71 \pm 0.08$ & $-0.28 \pm 0.01$ \\ \midrule
\multirow{7}{*}{\textbf{Hard}} 
 & MPC & $83.10 \pm 1.47$ & $16.70 \pm 1.47$ & $0.20 \pm 0.00$ &  - & $19.24 \pm 0.04$ & $-0.38 \pm 0.04$ \\
 & DRL-ORMOC & $90.09 \pm 0.86$ & $9.78 \pm 0.81$ & $0.13 \pm 0.13$ &  $138.90 \pm 0.20$ & $27.33 \pm 0.14$ & $-0.21 \pm 0.03$ \\
 & No Encoder & $97.41 \pm 0.68$ & $0.13 \pm 0.13$ & $2.46 \pm 0.57$ & $161.15 \pm 2.08$ & $28.96 \pm 0.07$ & $-0.24 \pm 0.01$ \\
 & DAROM-MLP & $0.00 \pm 0.00$ & $0.00 \pm 0.00$ & $0.00 \pm 0.00$ & $-196.62 \pm 0.27$ & $0.14 \pm 0.00$ & \bm{$0.01 \pm 0.00$} \\
 & DAROM-GRU w/o SC & $97.60 \pm 0.23$ & $2.40 \pm 0.23$ & \bm{$0.00 \pm 0.00$} & $174.26 \pm 0.63$ & $29.35 \pm 0.06$ & $-0.02 \pm 0.01$ \\
 & \textbf{DAROM-GRU} & \bm{$99.80 \pm 0.23$} & \bm{$0.00 \pm 0.00$} & $0.20 \pm 0.23$ & \bm{$179.68 \pm 0.51$} & \bm{$29.53 \pm 0.00$} & $-0.02 \pm 0.00$ \\
 & DAROM-Transformer & $92.08 \pm 1.28$ & $0.07 \pm 0.13$ & $7.85 \pm 1.16$ & $158.01 \pm 2.77$ & $29.35 \pm 0.03$ & $-0.37 \pm 0.01$ \\ \bottomrule
\end{tabular}
\begin{tablenotes}[flushleft]
\footnotesize
\item SC denotes safety controller.
\end{tablenotes}
\end{threeparttable}
\end{table*}

\begin{table*}[!h]
\centering
\caption{Ablation study. We compare the full DAROM-GRU architecture against variants lacking specific augmented state components and linear state predictor.}
\label{tab:ablation_results}
\begin{tabular}{@{}lcccc@{}}
\toprule
\textbf{Ablation Variant} & \textbf{Success Rate (\%)} & \textbf{Collision Rate (\%)} & \textbf{No Merge Rate (\%)} & \textbf{Avg. Return} \\ \midrule
Delayed State Only & $41.25 \pm 2.27$ & $56.75 \pm 2.27$ & {$2.00 \pm 0.00$} & $-0.83 \pm 5.39$ \\
Delayed State + Delay Magnitude & $0.00 \pm 0.00$ & $100.00 \pm 0.00$ & \bm{$0.00 \pm 0.00$} & $-73.47 \pm 0.77$ \\
Delayed State + Action Buffer & $0.00 \pm 0.00$ & $100.00 \pm 0.00$ & \bm{$0.00 \pm 0.00$} & $-67.84 \pm 1.52$ \\
Linear State Predictor & $94.94 \pm 2.91$ & $4.79 \pm 3.04$ & $0.00 \pm 0.00$ & $132.78 \pm 6.02$ \\
\textbf{Full Augmented State} & \bm{$99.80 \pm 0.23$} & \bm{$0.00 \pm 0.00$} & $0.20 \pm 0.23$ & \bm{$179.68 \pm 0.51$} \\ \bottomrule
\end{tabular}
\end{table*}

\begin{table*}[!h]
    \centering
    \caption{Performance comparison across delay profiles (Hard mode). DAROM-GRU architecture is trained under uniform distribution and evaluated under all five distributions.}
    \label{tab:delay_profile_results}
    \begin{tabular}{@{}lcccc@{}}
    \toprule
    \textbf{Delay Profile} & \textbf{Success Rate (\%)} & \textbf{Collision Rate (\%)} & \textbf{No Merge Rate (\%)} & \textbf{Avg. Return} \\ \midrule
Uniform & {$99.80 \pm 0.23$} & \bm{$0.00 \pm 0.00$} & $0.20 \pm 0.23$ &
  $179.68 \pm 0.51$ \\
      Bimodal & \bm{$100.00 \pm 0.00$} & \bm{$0.00 \pm 0.00$} & \bm{$0.00 \pm
  0.00$} & $180.03 \pm 0.06$ \\
      Bursty & \bm{$100.00 \pm 0.00$} & \bm{$0.00 \pm 0.00$} & \bm{$0.00 \pm
  0.00$} & $180.01 \pm 0.03$ \\
      Triangular & \bm{$100.00 \pm 0.00$} & \bm{$0.00 \pm 0.00$} & \bm{$0.00 \pm
  0.00$} & $180.01 \pm 0.07$ \\
      Exponential & \bm{$100.00 \pm 0.00$} & \bm{$0.00 \pm 0.00$} & \bm{$0.00 \pm
  0.00$} & \bm{$180.04 \pm 0.04$} \\ \bottomrule
    \end{tabular}
\end{table*}

\subsubsection{Comparative Performance Analysis}
The results in Table \ref{tab:main_results} show that the proposed DAROM-GRU architecture significantly outperforms all baselines in high density traffic scenarios. In the Hard mode, where traffic density is highest, the MPC baseline achieves only an 83.10\% success rate with a high collision rate of 16.70\%, highlighting its inability to account for stochastic latency in its prediction horizon. Similarly, the DRL-ORMOC baseline degrades to a 90.09\% success rate and a 9.78\% collision rate.

In contrast, {DAROM-GRU} maintains a near-perfect success rate of 99.80\% with 0.00\% collisions. While the ``No Encoder'' baseline achieves a comparable success rate of 97.41\%, it fails to guarantee safety, resulting in a 2.46\% collision rate. This indicates that while a standard policy can learn to merge, it cannot effectively avoid accidents under variable delays without the explicit state augmentation embeddings provided by the Delay-Aware Encoder.

\subsubsection{Efficacy of the Delay-Aware Encoder}
The choice of encoder architecture is critical for solving the RDMDP. The GRU-based encoder proved most effective at capturing the temporal dependencies of the action history. The MLP encoder failed completely in the Hard scenario (0.00\% success). This suggests that the feed-forward architecture of the MLP encoder is insufficient for the network to disentangle the causal effects of past actions over variable delay periods, as it lacks the recurrent inductive bias necessary to model temporal sequences effectively (at least in the Hard mode). The Transformer variant performed well (92.08\% success) but did not match the consistency of the GRU, likely because its larger capacity made it more prone to undertraining under our training budget.

\subsubsection{Safety Controller Impact}
The integration of the safety controller provides a final layer of robustness. Comparing DAROM-GRU with its ablated version ``DAROM-GRU w/o SC'' in Table \ref{tab:main_results} (Hard mode) reveals that the safety controller is crucial for eliminating tail-end risks. Without the SC, the agent achieves a 97.60\% success rate but suffers a 2.40\% collision rate. The inclusion of the safety controller eliminates these accidents entirely, reducing the collision rate to 0.00\% without a reduction in average velocity.

\subsubsection{Ablation Study}
Table \ref{tab:ablation_results} confirms the necessity of the full augmented state ($o_t, u_t, \omega_t$) for recovering the Markov property. In the Hard scenario, variants using only the Delayed State + Delay Magnitude or only the Delayed State + Action Buffer both yielded 0.00\% success rates. Furthermore, using the Delayed State alone resulted in a poor success rate of 41.25\%. These results confirm that the agent can only infer the true latent state when it processes \textit{both} the action history $u_t$ and the delay magnitude $\omega_t$ jointly. Additionally, the Linear State Predictor, which propagates surrounding vehicle positions forward using a constant-velocity model, achieves a 94.94\% success rate but retains a 4.79\% collision rate, demonstrating that physics-based state extrapolation alone is insufficient for safe merging under stochastic delays without the learned context provided by the full augmented state.

\subsubsection{Generalization Across Delay Profiles}
Table~\ref{tab:delay_profile_results}
   evaluates DAROM-GRU across five delay distributions unseen during training. The agent achieves perfect
   or near-perfect performance across all profiles. Notably, the uniform          
  distribution, which is also the training distribution, exhibits the only non-zero no-merge  
  rate ($0.20\% \pm 0.23\%$), which is statistically indistinguishable from zero  
  given the overlapping confidence interval. This is consistent with the          
  maximum-entropy nature of the uniform distribution: unlike bimodal, bursty,
  triangular, or exponential profiles, the uniform distribution lacks exploitable
  temporal structure, occasionally producing worst-case gap sequences that cause
  the conservative policy to timeout rather than merge. The failure mode is
  exclusively timeout with zero collisions, confirming that the safety-first
  behavior induced by training generalizes robustly. These results demonstrate
  that DAROM-GRU learns a delay-robust policy that transfers across heterogeneous
  real-world latency profiles without retraining.

%% file: 06-Conclusion/conclusion.tex
In this work, we proposed DAROM, a delay-aware reinforcement learning framework for on-ramp merging under V2I stochastic communication latency. By modeling the problem as an RDMDP and introducing a Delay-Aware Encoder, we enabled the agent to recover the system's latent state from delayed observations and action histories.

Extensive experiments demonstrated that DAROM-GRU achieves superior robustness compared to MPC and standard RL baselines, maintaining a 99.8\% success rate in high-density traffic with random delays up to 2.0 seconds. The results highlight the critical importance of jointly processing action history and delay magnitude to mitigate the loss of the Markov~property.

All experiments were conducted in SUMO with simulation parameters derived from NGSIM. While this improves realism relative to purely synthetic settings, a nontrivial sim-to-real gap may remain due to factors not captured by our simulator and data-driven calibration, such as sensor noise and missed detections in infrastructure perception, communication packet loss and nonstationary latency. These limitations are left for future work, including more realistic perception models, robustness evaluation under perturbed observation channels, and validation with field data.